\documentclass[11pt,a4paper]{article}
\usepackage[hyperref]{acl2018}
\usepackage{times}
\usepackage{latexsym}
\usepackage{graphicx}
\usepackage{caption}
\usepackage{subcaption}
\usepackage{url}
\usepackage{cleveref}
\usepackage{amsmath}
\usepackage{bm}
\usepackage{url}
\usepackage{multirow}

\newcommand{\z}[1]{\bm{z_{#1}}}
\newcommand{\s}[1]{\bm{s_{#1}}}
\newcommand{\w}[1]{\bm{w_{#1}}}
\newcommand{\bb}[1]{\bm{b_{#1}}}
\newcommand{\W}[1]{\bm{W}_{#1}}
\newcommand{\bS}[1]{\bm{S_{#1}}}
\newcommand{\m}[1]{\bm{m^{(#1)}}}
\newcommand{\M}[1]{\bm{M^{(#1)}}}
\newcommand{\bc}[1]{\bm{c^{(#1)}}}
\newcommand{\C}[1]{\bm{C^{(#1)}}}
\newcommand{\q}{\bm{q}}
\newcommand{\p}[2]{\bm{p_{#1}^{(#2)}}}
\newcommand{\f}{\bm{f}}

\aclfinalcopy

\title{Think Visually: Question Answering through Virtual Imagery}

\author{Ankit Goyal \qquad Jian Wang \qquad Jia Deng \\
  Computer Science and Engineering \\
  University of Michigan, Ann Arbor \\
  {\tt \{ankgoyal, jianwolf, jiadeng\}@umich.edu} \\}

\date{}

\begin{document}

\maketitle

\begin{abstract}
In this paper, we study the problem of geometric reasoning in the context of question-answering. We introduce Dynamic Spatial Memory Network (DSMN), a new deep network architecture designed for answering questions that admit latent visual representations. DSMN learns to generate and reason over such representations. Further, we propose two synthetic benchmarks, FloorPlanQA and ShapeIntersection, to evaluate the geometric reasoning capability of QA systems. Experimental results validate the effectiveness of our proposed DSMN for visual thinking tasks\footnote{\label{code} Code and datasets: \url{https://github.com/umich-vl/think_visually}}.
\end{abstract}

\section{Introduction}
The ability to reason is a hallmark of intelligence and a requirement for building question-answering (QA) systems. In AI research, reasoning has been strongly associated with logic and symbol manipulation, as epitomized by work in automated theorem proving~\citep{fitting2012first}. But for humans, reasoning involves not only symbols and logic, but also images and shapes. Einstein famously wrote: ``The psychical entities which seem to serve as elements in thought are certain signs and more or less clear images which can be `voluntarily' reproduced and combined... Conventional words or other signs have to be sought for laboriously only in a secondary state...'' And the history of science abounds with discoveries from visual thinking, from the Benzene ring to the structure of DNA~\citep{pinker2003language}.

There are also plenty of ordinary examples of human visual thinking. Consider a square room with a door in the middle of its southern wall. Suppose you are standing in the room such that the eastern wall of the room is behind you. Where is the door with respect to you? The answer is `to your left.' Note that in this case both the question and answer are just text. But in order to answer the question, it is natural to construct a mental picture of the room and use it in the process of reasoning. Similar to humans, the ability to `think visually' is desirable for AI agents like household robots. An example could be to construct a rough map and navigation plan for an unknown environment from verbal descriptions and instructions.

In this paper, we investigate how to model geometric reasoning (a form of visual reasoning) using deep neural networks (DNN). Specifically, we address the task of answering questions through geometric reasoning---both the question and answer are expressed in symbols or words, but a geometric representation is created and used as part of the reasoning process. 

In order to focus on geometric reasoning, we do away with natural language by designing two synthetic QA datasets, FloorPlanQA and ShapeIntersection. In FloorPlanQA, we provide the blueprint of a house in words and ask questions about location and orientation of objects in it. For ShapeIntersection, we give a symbolic representation of various shapes and ask how many places they intersect. In both datasets, a reference visual representation is provided for each sample.

Further, we propose Dynamic Spatial Memory Network (DSMN), a novel DNN that uses virtual imagery for QA. DSMN is similar to existing memory networks~\citep{kumar2016, sukhbaatar2015, henaff2016tracking} in that it uses vector embeddings of questions and memory modules to perform reasoning. The main novelty of DSMN is that it creates virtual images for the input question and uses a spatial memory to aid the reasoning process.

We show through experiments that with the aid of an internal visual representation and a spatial memory, DSMN outperforms strong baselines on both FloorPlanQA and ShapeIntersection. We also demonstrate that explicitly learning to create visual representations further improves performance. Finally, we show that DSMN is substantially better than the baselines even when visual supervision is provided for only a small proportion of the samples.

It's important to note that our proposed datasets consist of synthetic questions as opposed to natural texts. Such a setup allows us to sidestep difficulties in parsing natural language and instead focus on geometric reasoning. However, synthetic data lacks the complexity and diversity of natural text. For example, spatial terms used in natural language have various ambiguities that need to resolved by context (e.g. how far is "far" and whether "to the left" is relative to the speaker or the listener)~\citep{shariff1998natural, landau1993whence}, but our synthetic data lacks such complexities. Therefore, our method and results do not automatically generalize to real-life tasks involving natural language. Additional research is needed to extend and validate our approach on natural data.

Our contributions are three-fold: First, we present Dynamic Spatial Memory Network (DSMN), a novel DNN that performs geometric reasoning for QA. Second, we introduce two synthetic datasets that evaluate a system's visual thinking ability. Third, we demonstrate that on synthetic data, DSMN achieves superior performance for answering questions that require visual thinking.

\section{Related Work}
\noindent \textbf{Natural language datasets for QA:}
Several natural language QA datasets have been proposed to test AI systems on various reasoning abilities~\citep{levesque2011winograd, richardson2013mctest}. Our work differs from them in two key aspects: first, we use synthetic data instead of natural data; and second, we specialize in geometrical reasoning instead of general language understanding. Using synthetic data helps us simplify language parsing and thereby focus on geometric reasoning. However, additional research is necessary to generalize our work to natural data.

\noindent \textbf{Synthetic datasets for QA:}
Recently, synthetic datasets for QA are also becoming crucial in AI. In particular, bAbI~\citep{weston2015} has driven the development of several recent DNN-based QA systems~\citep{kumar2016, sukhbaatar2015, henaff2016tracking}. bAbI consists of 20 tasks to evaluate different reasoning abilities. Two tasks, Positional Reasoning (PR) and Path Finding (PF), are related to geometric reasoning. However, each Positional Reasoning question contains only two sentences, and can be solved through simple logical deduction such as `A is left of B implies B is right of A'. Similarly, Path Finding involves a search problem that requires simple spatial deductions such as `A is east of B implies B is west of A'. In contrast, the questions in our datasets involve longer descriptions, more entities, and more relations; they are thus harder to answer with simple deductions. We also provide reference visual representation for each sample, which is not available in bAbI.

\noindent \textbf{Mental Imagery and Visual Reasoning:}
The importance of visual reasoning has been long recognized in AI~\citep{forbus1991qualitative, lathrop2007towards}. Prior works in NLP~\citep{seo2015,lin2015} have also studied visual reasoning. Our work is different from them as we use synthetic language instead of natural language. Our synthetic language is easier to parse, allowing our evaluation to mainly reflect the performance of geometric reasoning. On the other hand, while our method and conclusions can potentially apply to natural text, this remains to be validated and involves nontrivial future work. There are other differences to prior works as well. Specifically,~\citep{seo2015} combined information from textual questions and diagrams to build a model for solving SAT geometry questions. However, our task is different as diagrams are not provided as part of the input, but are generated from the words/symbols themselves. Also,~\citep{lin2015} take advantage of synthetic images to gather semantic common sense knowledge (visual common sense) and use it to perform fill-in-the-blank (FITB) and visual paraphrasing tasks. Similar to us, they also form `mental images'. However, there are two differences (apart from natural vs synthetic language): first, their benchmark tests higher level semantic knowledge (like ``Mike is having lunch when he sees a bear.'' $\implies$ ``Mike tries to hide.''), while ours is more focused on geometric reasoning. Second, their model is based on hand-crafted features while we use a DNN.

\noindent \textbf{Spatial language for Human-Robot Interaction:}
Our work is also related to prior work on making robots understand spatial commands  (e.g.\ ``put that box here'', ``move closer to the box'') and complete tasks such as navigation and assembly. Earlier work~\citep{muller2000coarse, gribble1998integrating, zelek1997human} in this domain used template-based commands, whereas more recent work~\citep{skubic2004spatial} tried to make the commands more natural. This line of work differs from ours in that the robot has visual perception of its environment that allows grounding of the textual commands, whereas in our case the agent has no visual perception, and an environment needs to be imagined.

\noindent \textbf{Image Generation:}
Our work is related to image generation using DNNs which has a large body of literature, with diverse approaches~\citep{reed2016generative, gregor2015draw}. We also generate an image from the input. But in our task, image generation is in the service of reasoning rather than an end goal in itself---as a result, photorealism or artistic style of generated images is irrelevant and not considered.

\noindent \textbf{Visual Question Answering:}
Our work is also related to visual QA (VQA)~\citep{johnson2016, antol2015, lu2016hierarchical}. Our task is different from VQA because our questions are in terms of words/symbols whereas in VQA the questions are visual, consisting of both text descriptions and images. The images involved in our task are internal and virtual, and are not part of the input or output.

\noindent \textbf{Memory and Attention:}
Memory and attention have been increasingly incorporated into DNNs, especially for tasks involving algorithmic inference and/or natural language~\citep{graves2014neural,vaswani2017attention}. For QA tasks, memory and attention play an important role in state-of-the-art (SOTA) approaches.~\citep{sukhbaatar2015} introduced End-To-End Memory Network (MemN2N), a DNN with memory and recurrent attention mechanism, which can be trained end-to-end for diverse tasks like textual QA and language modeling. Concurrently,~\cite{kumar2016} introduced Dynamic Memory Network (DMN), which also uses attention and memory.~\cite{xiong2016} proposed DMN+, with several improvements over the previous version of DMN and achieved SOTA results on VQA~\citep{antol2015} and bAbI~\citep{weston2015}. Our proposed DSMN is a strict generalization of DMN+ (see Sec.~\ref{generalize_dmn}). On removing the images and spatial memory from DSMN, it reduces to DMN+. Recently~\cite{gupta2017cognitive} also used spatial memory in their deep learning system, but for visual navigation. We are using spatial memory for QA.
\begin{figure}
  \centering
  \includegraphics[width=0.5\textwidth]{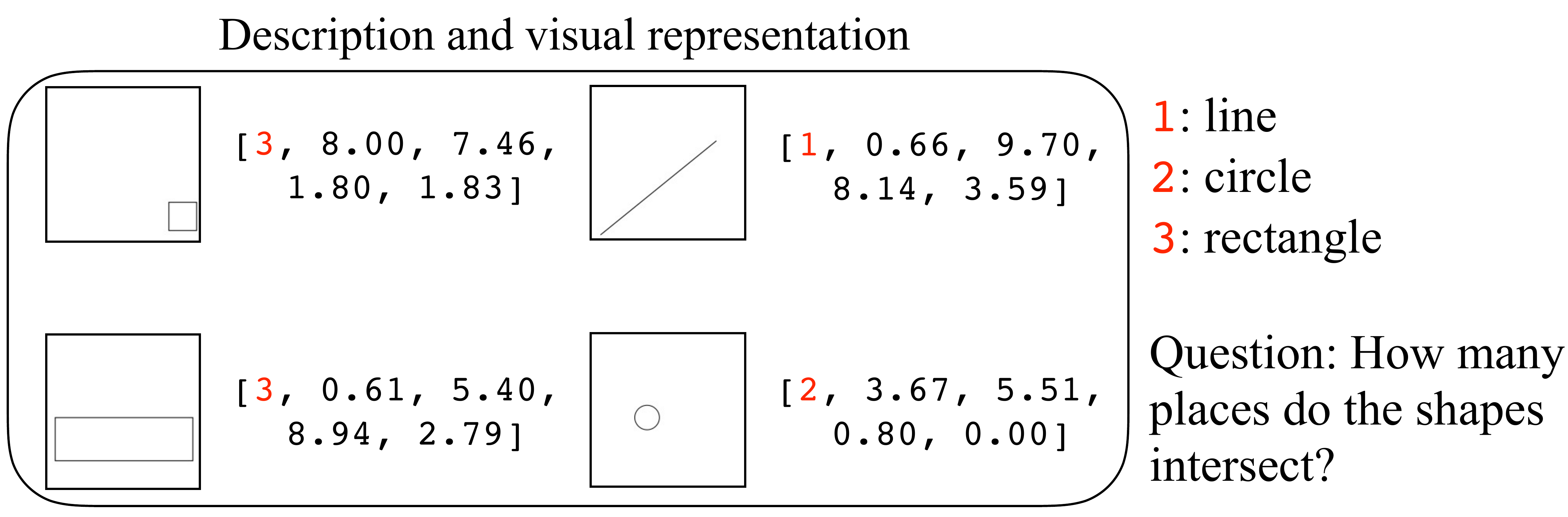}
  \vspace{-2mm}
  \caption{An example in the ShapeIntersection dataset.}
  \label{fig:shape_inter}
  \vspace{-2mm}
\end{figure}
\begin{table*}
  \centering
  \begin{center}
    \footnotesize
    \begin{tabular}{l||p{13.7cm}}
    Component   & Template \\
    \hline \hline
    \multirow{2}{1cm}{House door}  & The house door is in the middle of the \{\textit{nr, sr, er, wr}\} wall of the house. \\
                                 & The house door is located in the \{\textit{n-er, s-er, n-wr, s-wr, n-er, s-er, n-wr, s-wr}\} side of the house, such that it opens towards \{\textit{n, s, e, w}\}. \\
    \hline                          
    \multirow{4}{1cm}{Room door}   & The door for this room is in the middle of its \{\textit{nr, sr, er, wr}\} wall. \\
                                 & This room's door is in the middle of its \{\textit{nr, sr, er, wr}\} wall. \\
                                 & The door for this room is located in its \{\textit{n-er, s-er, n-wr, s-wr, n-er, s-er, n-wr, s-wr}\} side, such that it opens towards \{\textit{n, s, e, w}\}. \\
                                 & This room's door is located in its \{\textit{n-er, s-er, n-wr, s-wr, n-er, s-er, n-wr, s-wr}\} side, such that it opens towards \{\textit{n, s, e, w}\}. \\
    \hline
    \multirow{2}{1cm}{Small room}  & Room \{1, 2, 3\} is small in size and it is located in the \{\textit{n, s, e, w, c, n-e, s-e, n-w, s-w}\} of the house. \\
                                 & Room \{1, 2, 3\} is located in the \{\textit{n, s, e, w, c, n-e, s-e, n-w, s-w}\} of the house and is small in size. \\
    \hline
    \multirow{2}{1cm}{Medium room} & Room \{1, 2, 3\} is medium in size and it extends from the \{\textit{n, s, e, w, c, n-e, s-e, n-w, s-w}\} to the \{\textit{n, s, e, w, c, n-e, s-e, n-w, s-w}\} of the house. \\
                                 & Room \{1, 2, 3\} extends from the \{\textit{n, s, e, w, c, n-e, s-e, n-w, s-w}\} to the \{\textit{n, s, e, w, c, n-e, s-e, n-w, s-w}\} of the house and is medium in size. \\
    \hline
    \multirow{2}{1cm}{Large room}  & Room \{1, 2, 3\} is large in size and it stretches along the \{\textit{n-s, e-w}\}direction in the \{\textit{n, s, e, w, c}\} of the house. \\
                                 & Room \{1, 2, 3\} stretches along the \{\textit{n-s, e-w}\} direction in the \{\textit{n, s, e, w, c}\} of the house and is large in size. \\
    \hline
    \multirow{4}{*}{Object}      & A \{\textit{cu, cd, sp, co}\} is located in the middle of the \{\textit{nr, sr, er, wr}\} part of the house. \\
                                 & A \{\textit{cu, cd, sp, co}\} is located in the \{\textit{n-er, s-er, n-wr, s-wr, n-er, s-er, n-wr, s-wr, cr}\} part of the house. \\
                                 & A \{\textit{cu, cd, sp, co}\} is located in the middle of the \{\textit{nr, sr, er, wr}\} part of this room. \\
                                 & A \{\textit{cu, cd, sp, co}\} is located in the \{\textit{n-er, s-er, n-wr, s-wr, n-er, s-er, n-wr, s-wr, cr}\} part of this room. \\
    \hline
    \end{tabular}
  \end{center}
  \vspace{-2mm}
  \caption{Templates used by the description generator for FloorPlanQA. For compactness we used the following notations, \textit{n} - north, \textit{s} - south, \textit{e} - east, \textit{w} - west, \textit{c} - center, \textit{nr} - northern, \textit{sr} - southern, \textit{er} - eastern, \textit{wr} - western, \textit{cr} - central, \textit{cu} - cube, \textit{cd} - cuboid, \textit{sp} - sphere and \textit{co} - cone.}
  \label{tab:template}
  \vspace{-2mm}
\end{table*}
\section{Datasets}
\label{sec:datasets}
We introduce two synthetically-generated QA datasets to evaluate a system's goemetrical reasoning ability: FloorPlanQA and ShapeIntersection. These datasets are not meant to test natural language understanding, but instead focus on geometrical reasoning. Owing to their synthetic nature, they are easy to parse, but nevertheless they are still challenging for DNNs like DMN+~\citep{xiong2016} and MemN2N~\citep{sukhbaatar2015} that achieved SOTA results on existing QA datasets (see Table~\ref{tab:all}).

The proposed datasets are similar in spirit to bAbI~\citep{weston2015}, which is also synthetic. In spite of its synthetic nature, bAbI has proved to be a crucial benchmark for the development of new models like MemN2N, DMN+, variants of which have proved successful in various natural domains~\citep{kumar2016, perez2016dialog}. Our proposed dataset is first to explicitly test `visual thinking', and its synthetic nature helps us avoid the expensive and tedious task of collecting human annotations. Meanwhile, it is important to note that conclusions drawn from synthetic data do not automatically translate to natural data, and methods developed on synthetic benchmarks need additional validation on natural domains. 

The proposed datasets also contain visual representations of the questions. Each of them has 38,400 questions, evenly split into a training set, a validation set and a test set (12,800 each).

\noindent \textbf{FloorPlanQA:}
Each sample in FloorPlanQA involves the layout of a house that has multiple rooms (max 3). The rooms are either small, medium or large. All the rooms and the house have a door. Additionally, each room and empty-space in the house (i.e. the space in the house that is not part of any room) might also contain an object (either a cube, cuboid, sphere, or cone). 

\begin{figure*}
  \begin{subfigure}{.62\textwidth}
    \includegraphics[width=\textwidth]{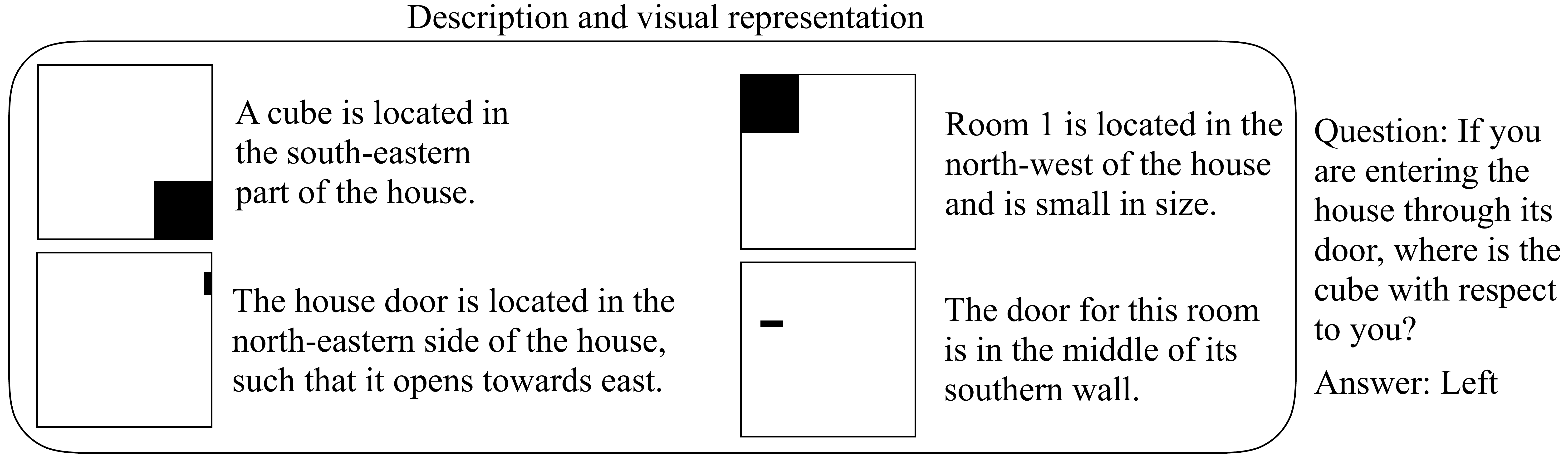}
  \end{subfigure}
  \begin{subfigure}{.38\textwidth}
    \begin{center}
      \footnotesize
      \begin{tabular}{l||c}
      vocabulary size                      & 66    \\
      \# unique sentences                  & 264   \\
      \# unique descriptions               & 38093 \\
      \# unique questions                  & 32    \\
      \# unique question-description pairs & 38228 \\
      Avg. \# words per sentence           & 15    \\
      Avg. \# sentences per description    & 6.61  \\
      \end{tabular}
    \end{center}
  \end{subfigure}
  \vspace{-2mm}
  \caption{An example and characteristics of FloorPlanQA (when considering all the 38,400 samples i.e. training, validation and test sets combined).}
  \label{fig:floorplan_qa}
  \vspace{-2mm}
\end{figure*}

Each sample has four components, \textit{a description, a question, an answer}, and \textit{a visual representation}. Each sentence in the description describes either a room, a door or an object. A question is of the following template: \textit{Suppose you are entering the \{house, room 1, room 2, room 3\}, where is the \{house door, room 1 door, room 2 door, room 3 door, cube, cuboid, sphere, cone\} with respect to you?}. The answer is either of \textit{left, right, front, } or \textit{back}. Other characteristics of FloorPlanQA are summarized in Fig.~\ref{fig:floorplan_qa}.

The visual representation of a sample consists of an ordered set of image channels, one per sentence in the description. An image channel pictorially represents the location and/or orientation of the described item (room, door, object) w.r.t. the house. An example is shown in Fig.~\ref{fig:floorplan_qa}.

To generate samples for FloorPlanQA, we define a probabilistic generative process which produces tree structures representing layouts of houses, similar to scene graphs used in computer graphics. The root node of a tree represents an entire house, and the leaf nodes represent rooms. We use a description and visual generator to produce respectively the description and visual representation from the tree structure. The templates used by the description generator are described in Table~\ref{tab:template}. Furthermore, the order of sentences in a description is randomized while making sure that the description still makes sense. For example, in some sample, the description of room 1 can appear before that of the house-door, while in another sample, it could be reversed. Similarly, for a room, the sentence describing the room's door could appear before or after the sentence describing the object in the room (if the room contains one). We perform rejection sampling to ensure that all the answers are equally likely, and thus removing bias. 

\noindent \textbf{ShapeIntersection:}
As the name suggests, ShapeIntersection is concerned with counting the number of intersection points between shapes. In this dataset, the description consists of symbols representing various shapes, and the question is always ``how many points of intersection are there among these shapes?''

There are three types of shapes in ShapeIntersection: rectangles, circles, and lines. The description of shapes is provided in the form of a sequence of 1D vectors, each vector representing one shape. A vector in ShapeIntersection is analogous to a sentence in FloorPlanQA. Hence, for ShapeIntersection, the term `sentence' actually refers to a vector. Each sentence describing a shape consists of 5 real numbers. The first number stands for the type of shape: 1 - line, 2 - circle, and 3 - rectangle. The subsequent four numbers specify the size and location of the shape. For example, in case of a rectangle, they represent its height, its width, and coordinates of its bottom-left corner. Note that one can also describe the shapes using a sentence, e.g. ``there is a rectangle at (5, 5), with a height of 2\,cm and width of 8\,cm.'' However, as our focus is to evaluate `visual thinking', we work directly with the symbolic encoding.

In a given description, there are 6.5 shapes on average, and at most 6 lines, 3 rectangles and 3 circles. All the shapes in the dataset are unique and lie on a $10 \times 10$ canvas. While generating the dataset, we do rejection sampling to ensure that the number of intersections is uniformly distributed from 0 to the maximum possible number of intersections, regardless of the number of lines, rectangles, and circles. This ensures that the number of intersections cannot be estimated from the number of lines, circles or rectangles.

Similar to FloorPlanQA, the visual representation for a sample in this dataset is an ordered set of image channels. Each channel is associated with a sentence, and it plots the described shape. An example is shown in~\Cref{fig:shape_inter}. 
~
\section{Dynamic Spatial Memory Network}
We propose Dynamic Spatial Memory Network (DSMN), a novel DNN designed for QA with geometric reasoning. What differentiates DSMN from other QA DNNs is that it forms an internal visual representation of the input. It then uses a spatial memory to reason over this visual representation.

\begin{figure*}
  \centering
  \includegraphics[width=0.9\textwidth]{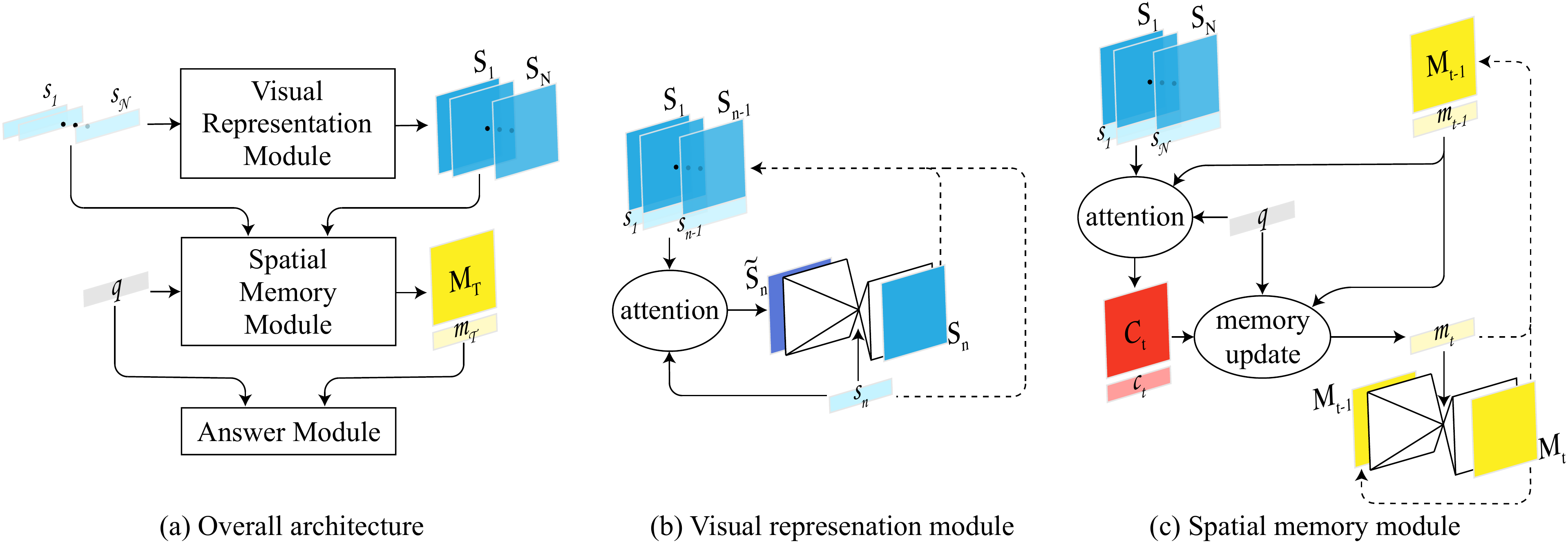}
  \vspace{-2mm}
  \caption{The architecture of the proposed Dynamic Spatial Memory Network (DSMN).}
  \centering
  \label{dsmn}
  \vspace{-2mm}
\end{figure*}

A DSMN can be divided into five modules: \textit {the input module, visual representation module, question module, spatial memory module}, and \textit{answer module}. The input module generates an embedding for each sentence in the description. The visual representation module uses these embeddings to produce an intermediate visual representation for each sentence. In parallel, the question module produces an embedding for the question. The spatial memory module then goes over the question embedding, the sentence embeddings, and the visual representation multiple times to update the spatial memory. Finally, the answer module uses the spatial memory to output the answer. Fig.~\ref{dsmn} illustrates the overall architecture of DSMN.

\noindent \textbf{Input Module:} 
This module produces an embedding for each sentence in the description. It is therefore customized based on how the descriptions are provided in a dataset. Since the descriptions are in words for FloorPlanQA, a position encoding (PE) layer is used to produce the initial sentence embeddings. This is done to ensure a fair comparison with DMN+~\citep{xiong2016} and MemN2N~\citep{sukhbaatar2015}, which also use a PE layer. A PE layer combines the word-embeddings to encode the position of words in a sentence (Please see ~\cite{sukhbaatar2015} for more information). For ShapeIntersection, the description is given as a sequence of vectors. Therefore, two FC layers (with ReLU in between) are used to obtain the initial sentence embeddings.

These initial sentence embeddings are then fed into a bidirectional Gated Recurrent Unit (GRU)~\citep{cho2014properties} to propagate the information across sentences. Let $\overrightarrow{\s{i}}$ and $\overleftarrow{\s{i}}$ be the respective output of the forward and backward GRU at $i^{th}$ step. Then, the final sentence embedding for the $i^{th}$ sentence is given by $\s{i} = \overrightarrow{\s{i}} + \overleftarrow{\s{i}}$.

\noindent \textbf{Question Module:} 
This module produces an embedding for the question. It is also customized to the dataset. For FloorPlanQA, the embeddings of the words in the question are fed to a GRU, and the final hidden state of the GRU is used as the question embedding. For ShapeIntersection, the question is always fixed, so we use an all-zero vector as the question embedding.

\noindent \textbf{Visual Representation Module:} 
This module generates a visual representation for each sentence in the description. It consists of two sub-components: an attention network and an encoder-decoder network. The attention network gathers information from previous sentences that is important to produce the visual representation for the current sentence. For example, suppose the current sentence describes the location of an object with respect to a room. Then in order to infer the location of the object with respect to the house, one needs the location of the room with respect to the house, which is described in some previous sentence.

The encoder-decoder network encodes the visual information gathered by the attention network, combines it with the current sentence embedding, and decodes the visual representation of the current sentence. An encoder ($En(.)$) takes an image as input and produces an embedding, while a decoder ($De(.)$) takes an embedding as input and produces an image. An encoder is composed of series of convolution layers and a decoder is composed of series of deconvolution layers.

Suppose we are currently processing the sentence $\s{t}$. This means we have already processed the sentences $\s{1}, \s{2}, \ldots, \s{t-1}$ and produced the corresponding visual representations $\bS{1}, \bS{2}, \ldots, \bS{t-1}$. We also add $\s{0}$ and $\bS{0}$, which are all-zero vectors to represent the null sentence. The attention network produces a scalar attention weight $a_{i}$ for the $i^{th}$ sentence which is given by $a_i = \text{Softmax}({\w{s}}^t \z{i} + b_{s})$ where $\z{i} = [|\s{i} - \s{t}|; \s{i} \circ \s{t}]$. Here, $\w{s}$ is a vector, $b_{s}$ is a scalar, $\circ$ represents element-wise multiplication, $|.|$ represents element-wise absolute value, and $[\bm{v1}; \bm{v2}]$ represents the concatenation of vectors $\bm{v1}$ and $\bm{v2}$. 

The gathered visual information is $\bar{\bS{t}} = \sum_{i=0}^{t-1} a_i \bS{i}$. It is fed into the encoder-decoder network. The visual representation for $\s{t}$ is given by $\bS{t} = De_s\Big(\big[\s{t}; En_s(\bar{\bS{t}})\big]\Big)$. The parameters of $En_s(.)$, $De_s()$, $\w{s}$, and $b_s$ are shared across multiple iterations.

In the proposed model, we make the simplifying assumption that the visual representation of the current sentence does not depend on future sentences. In other words, it can be completely determined from the previous sentences in the description. Both FloorPlanQA and ShapeIntersection satisfy this assumption.

\noindent \textbf{Spatial Memory Module:}
This module gathers relevant information from the description and updates memory accordingly. Similar to DMN+ and MemN2N, it collects information and updates memory multiple times to perform transitive reasoning. One iteration of information collection and memory update is referred as a `hop'.

The memory consists of two components: a 2D spatial memory and a tag vector. The 2D spatial memory can be thought of as a visual scratch pad on which the network `sketches' out the visual information. The tag vector is meant to represent what is `sketched' on the 2D spatial memory. For example, the network can sketch the location of room 1 on its 2D spatial memory, and store the fact that it has sketched room 1 in the tag vector.

As mentioned earlier, each step of the spatial memory module involves gathering of relevant information and updating of memory. Suppose we are in step $t$. Let $\M{t-1}$ represent the 2D spatial memory and $\m{t-1}$ represent the tag vector after step $t-1$. The network gathers the relevant information by calculating the attention value for each sentence based on the question and the current memory. For sentence $\s{i}$, the scalar attention value $g_i^{(t)}$ equal to $\text{Softmax}(\w{y}^t \p{i}{t} + b_{y})$, where $\p{i}{t}$ is given as
\begin{align}
  \p{i}{t} &= \big[|\m{t-1} - \s{i}|; \ \m{t-1} \circ \s{i}; \ |\q - \s{i}|; \nonumber \\
  & \q \circ \s{i}; \ {En}_{p_{1}}^{(t)}(|\M{t-1} - \bS{i}|); \nonumber\\
  & {En}_{p_{2}}^{(t)}(\M{t-1} \circ \bS{i}) \big]
\end{align}
$\M{0}$ and $\m{0}$ represent initial blank memory, and their elements are all zero. Then, gathered information is represented as a context tag vector, $\bc{t} = \text{AttGRU}({g_i}^{(t)}\s{i})$ and 2D context, $\C{t} = \sum_{i=0}^{n}{g_i}^{(t)}\bS{i}$. Please refer to ~\cite{xiong2016} for information about AttGRU(.). Finally, we use the 2D context and context tag vector to update the memory as follows:
\begin{align}
  \m{t} &= \text{ReLU}\Big({\W{m}}^{(t)}\big[\m{t-1}; \ \q; \ \bc{t}; \nonumber \\ & En_c(\C{t})\big] + {\bb{m}}^{(t)}\Big) \\
  \M{t} &= De_m^{(t)}\Big(\big[\m{t}; \ En_m^{(t)}(\M{t-1})\big]\Big)
\end{align}
~
\textbf{Answer Module:}
This module uses the final memory and question embedding to generate the output. The feature vector used for predicting the answer is given by $\f$, where $\M{T}$ and $\m{T}$ represent the final memory.
\begin{align}
\label{eqn:f_answer}
  \f = \big[En_f(\M{T}); \ \m{T}; \ \q \big]
\end{align}
To obtain the output, an FC layer is applied to $\f$ in case of regression, while the FC layer is followed by softmax in case of classification. To keep DSMN similar to DMN+, we apply a dropout layer on sentence encodings ($\s{i}$) and $\f$.
~
\subsection{DSMN as a strict generalization of DMN}
\label{generalize_dmn}
DSMN is a strict generalization of a DMN+. If we remove the visual representation of the input along with the 2D spatial memory, and just use vector representations with memory tags, then a DSMN reduces to DMN+. This ensures that comparison with DMN+ is fair.

\subsection{DSMN with or without intermediate visual supervision}
As described in previous sections, a DSMN forms an intermediate visual representation of the input. Therefore, if we have a `ground-truth' visual representation for the training data, we could use it to train our network better. This leads to two different ways for training a DSMN, one with intermediate visual supervision and one without it. Without intermediate visual supervision, we train the network in an end-to-end fashion by using a loss ($L_{w/o\_ vi}$) that compares the predicted answer with the ground truth. With intermediate visual supervision, we train our network using an additional visual representation loss ($L_{vi}$) that measures how close the generated visual representation is to the ground-truth representation. Thus, the loss used for training with intermediate supervision is given by $L_{w\_vi} = \lambda_{vi}L_{vi} + (1-\lambda_{vi})L_{w/o\_vi}$, where $\lambda_{vi}$ is a hyperparameter which can be tuned for each dataset. Note that in neither case do we need any visual input once the network is trained. During testing, the only input to the network is the description and question. 

Also note that we can provide intermediate visual supervision to DSMN even when the visual representations for only a portion of samples in the training data are available. This can be useful when obtaining visual representation is expensive and time-consuming.

\section{Experiments}
\noindent \textbf{Baselines:}
LSTM~\citep{hochreiter1997long} is a popular neural network for sequence processing tasks. We use two versions of LSTM-based baselines. LSTM-1 is a common version that is used as a baseline for textual QA~\citep{sukhbaatar2015, graves2016hybrid}. In LSTM-1, we concatenate all the sentences and the question to a single string. For FloorPlanQA, we do word embedding look-up, while for ShapeIntersection, we project each real number into higher dimension via a series of FC layers. The sequence of vectors is fed into an LSTM. The final output vector of the LSTM is then used for prediction.

We develop another version of LSTM that we call LSTM-2, in which the question is concatenated to the description. We use a two-level hierarchy to embed the description. We first extract an embedding for each sentence. For FloorPlanQA, we use an LSTM to get the sentence embeddings, and for ShapeIntersection, we use a series of FC layers. We then feed the sentence embeddings into an LSTM, whose output is used for prediction. 

Further, we compare our model to DMN+~\citep{xiong2016} and MemN2N~\citep{sukhbaatar2015}, which achieved state-of-the-art results on bAbI~\citep{weston2015}. In particular, we compare the 3-hop versions of DSMN, DMN+, and MemN2N.
~
\begin{table}[t]
  \centering
  \begin{subtable}{.5\textwidth}
    \centering
    \begin{center}
      \footnotesize
      \begin{tabular}{l||c|c}
              & FloorPlanQA          & ShapeIntersection \\
      MODEL   & (accuracy in \%) & (rmse) \\ 
      \hline \hline 
      LSTM-1  &     41.36      &      3.28     \\
      LSTM-2  &     50.69      &      2.99     \\
      MemN2N  &     45.92      &      3.51     \\
      DMN+    &     60.29      &      2.98     \\
      DSMN    &     68.01      &      2.84     \\
      DSMN*   & \textbf{97.73} & \textbf{2.14} \\
      \end{tabular}
    \end{center}
    \caption{The test set performance of different models on FloorPlanQA and ShapeIntersection. DSMN* refers to the model with intermediate supervision.}
    \label{tab:all}
  \end{subtable}
  ~ 
  \begin{subtable}{.5\textwidth}
    \centering
    \begin{center}
      \footnotesize
      \begin{tabular}{l||l|c}
                &                                   & FloorPlanQA \\
        MODEL   & $\f$ in Eqn.~$\ref{eqn:f_answer}$ & (accuracy in \%) \\
        \hline \hline 
        DSMN  & $\big[\m{T}; \ \q \big]$                & 67.65 \\
        DSMN  & $\big[En_f(\M{T}); \ \q \big]$          & 43.90 \\
        DSMN  & $\big[En_f(\M{T}); \ \m{T}; \ \q \big]$ & 68.12 \\
        DSMN* & $\big[\m{T}; \ \q \big]$                & 97.24 \\
        DSMN* & $\big[En_f(\M{T}); \ \q \big]$          & 95.17 \\
        DSMN* & $\big[En_f(\M{T}); \ \m{T}; \ \q \big]$ & 98.08 \\
      \end{tabular}
    \end{center}
    \caption{The validation set performances for the ablation study on the usefulness of tag ($\m{T}$) and 2D spatial memory ($\M{T}$) in the answer feature vector for $\f$.}
    \label{tab:answer}
  \end{subtable}
  ~
  \begin{subtable}{.5\textwidth}
    \centering
    \begin{center}
      \footnotesize
      \begin{tabular}{l||c}
                & FloorPlanQA \\
        MODEL   & (accuracy in \%) \\
        \hline \hline 
        1-Hop DSMN  & 63.32 \\
        2-Hop DSMN  & 65.59 \\
        3-Hop DSMN  & 68.12 \\
        1-Hop DSMN* & 90.09 \\
        2-Hop DSMN* & 97.45 \\
        3-Hop DSMN* & 98.08 \\
      \end{tabular}
    \end{center}
    \caption{The validation set performance for the ablation study on variation in performance with hops.}
    \label{tab:hops}
  \end{subtable}
  \vspace{-2mm}
  \caption{Experimental results showing comparison with baselines, and ablation study of DSMN}
  \vspace{-2mm}
\end{table}
~
~
\begin{figure}
  \centering
  \begin{subfigure}{0.45\textwidth}
    \includegraphics[width=\textwidth]{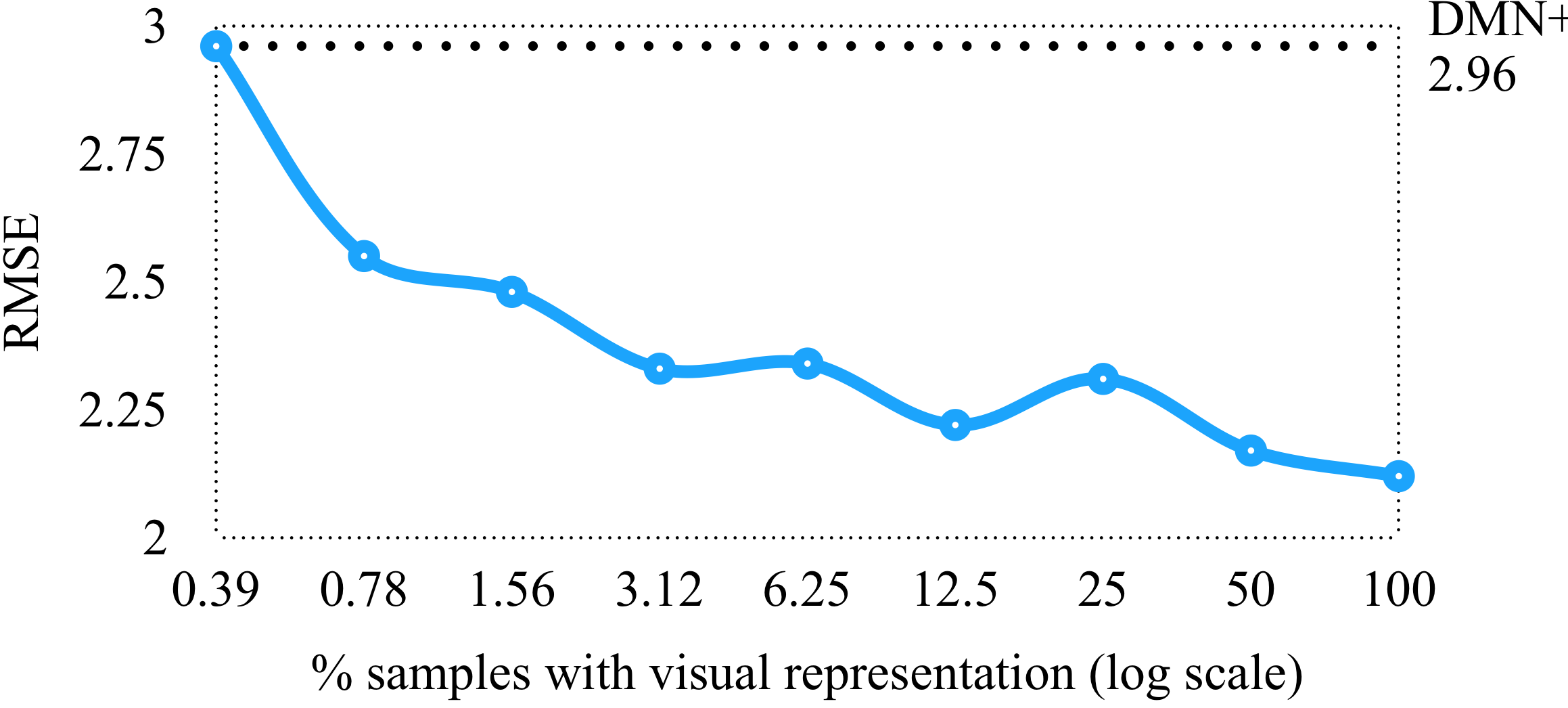}
    \caption{Test set rmse on ShapeIntersection.}
    \label{part1}
  \end{subfigure}
  ~ 
  \begin{subfigure}{0.45\textwidth}
    \includegraphics[width=\textwidth]{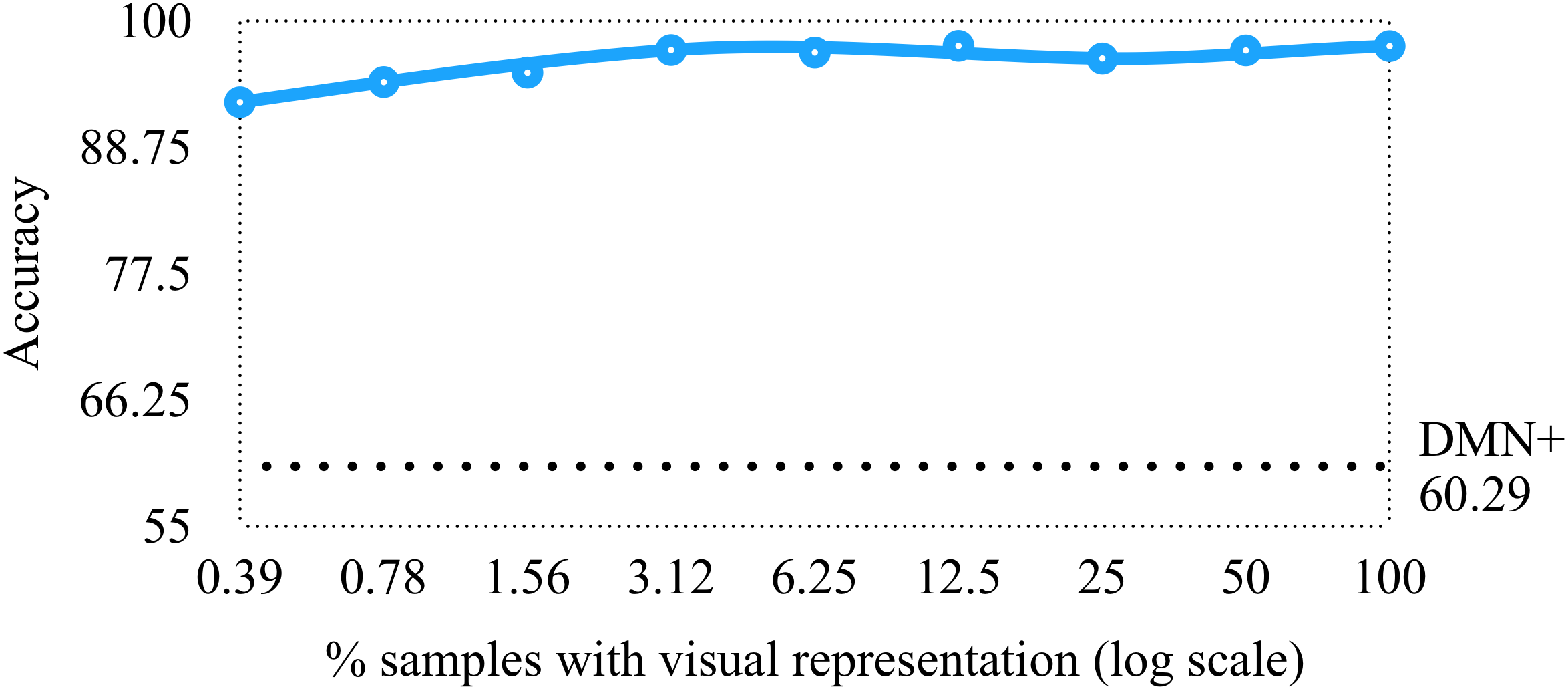}
    \caption{Test set accuracy on FloorPlanQA.}
    \label{part2}
  \end{subfigure}
  \vspace{-2mm}
  \caption{Performance of DSMN* with varying percentage of intermediate visual supervision.}
  \vspace{-2mm}
  \label{part}
\end{figure}
\begin{figure*}[t]
  \centering
  \includegraphics[width=0.88\textwidth]{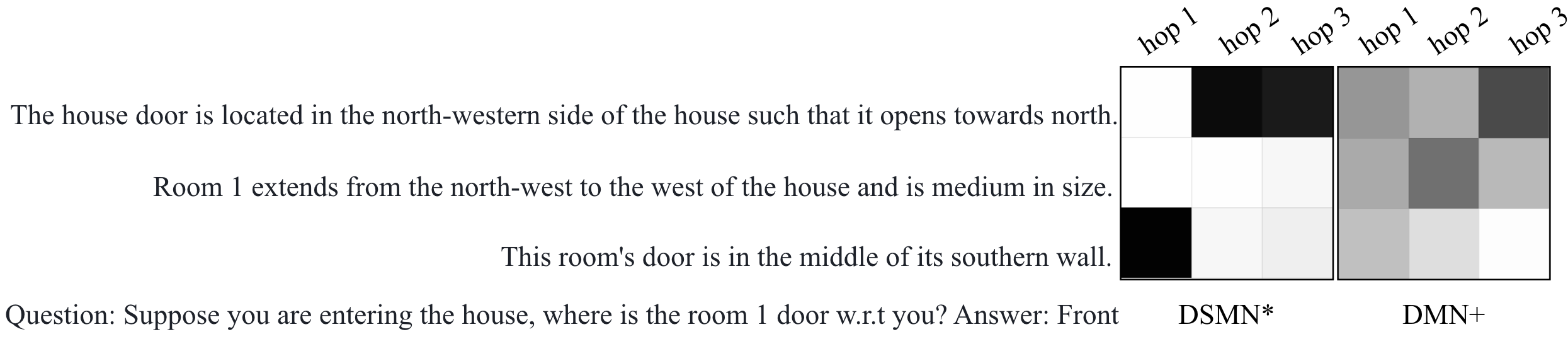}
  \vspace{-2mm}
  \caption{Attention values on each sentence during different memory `hops' for a sample from FloorPlanQA. Darker color indicates more attention. To answer, one needs the location of room 1's door and the house door. To infer the location of room 1's door, DSMN* directly jumps to sent. 3. Since DMN+ does not form a visual representation, it tries to infer the location of room 1's door w.r.t the house by finding the location of the room's door w.r.t the room (sent. 3) and the location of the room w.r.t the house (sent. 2). Both DSMN* and DMN+ use one hop to infer the location of the house door (sent. 1).}
  \label{visualization_fig}
  \vspace{-2mm}
\end{figure*}

\noindent \textbf{Training Details:}
We used ADAM~\citep{kingma2014adam} to train all models, and the learning rate for each model is tuned for each dataset. We tune the embedding size and $l_2$ regularization weight for each model and dataset pair separately. For reproducibility, the value of the best-tuned hyperparameters is mentioned in the appendix. As reported by~\citep{sukhbaatar2015, kumar2016, henaff2016tracking}, we also observe that the results of memory networks are unstable across multiple runs. Therefore for each hyperparameter choice, we run all the models 10 times and select the run with the best performance on the validation set. For FloorPlanQA, all models are trained up to a maximum of 1600 epochs, with early stopping after 80 epochs if the validation accuracy did not increase. The maximum number of epochs for ShapeIntersection is 800 epochs, with early stopping after 80 epochs. Additionally, we modify the input module and question module of DMN+ and MemN2N to be same as ours for the ShapeIntersection dataset.

For MemN2N, we use the publicly available implementation\footnote{\url{https://github.com/domluna/memn2n}} and train it exactly as all other models (same optimizer, total epochs, and early stopping criteria) for fairness. While the reported best result for MemN2N is on the version with position encoding, linear start training, and random-injection of time index noise~\citep{sukhbaatar2015}, the version we use has only position encoding. Note that the comparison is still meaningful because linear start training and time index noise are not used in DMN+ (and as a result, neither in our proposed DSMN).

\noindent \textbf{Results:}
The results for FloorPlanQA and ShapeIntersection are summarized in Table~\ref{tab:all}. For brevity, we will refer to the DSMN model trained without intermediate visual supervision as DSMN, and the one with intermediate visual supervision as DSMN*. We see that DSMN (i.e the one without intermediate supervision) outperforms DMN+, MemN2N and the LSTM baselines on both datasets. However, we consider DSMN to be only slightly better than DMN+ because both are observed to be unstable across multiple runs and so the gap between the two has a large variance. Finally, DSMN* outperforms all other approaches by a large margin on both datasets, which demonstrates the utility of visual supervision in proposed tasks. While the variation can be significant across runs, if we run each model 10 times and choose the best run, we observe consistent results. We visualized the intermediate visual representations, but when no visual supervision is provided, they were not interpretable (sometimes they looked like random noise, sometimes blank). In the case when visual supervision is provided, the intermediate visual representation is well-formed and similar to the ground-truth.

We further investigate how DSMN* performs when intermediate visual supervision is available for only a portion of training samples. As shown in Fig.~\ref{part}, DSMN* outperforms DMN+ by a large margin, even when intermediate visual supervision is provided for only $~1\%$ of the training samples. This can be useful when obtaining visual representations is expensive and time-consuming. One possible justification for why visual supervision (even in a small amount) helps a lot is that it constrains the high­-dimensional space of possible intermediate visual representations. With limited data and no explicit supervision, automatically learning these high-dimensional representations can be difficult.

Additonally, we performed ablation study (see Table~\ref{tab:answer}) on the usefulness of final memory tag vector ($\m{T}$) and 2D spatial memory ($\M{T}$) in the answer feature vector $\f$ (see Eqn.~\ref{eqn:f_answer}). We removed each of them one at a time, and retrained (with hyperparameter tuning) the DSMN and DSMN* models. Note that they are removed only from the final feature vector $\f$, and both of them are still coupled. The model with both tag and 2D spatial memory ($\f = \big[En_f(\M{T}); \m{T}; \q \big]$) performs slightly better than the only tag vector model ($\f = \big[\m{T}; \q \big]$). Also, as expected the only 2D spatial memory model ($\f = \big[En_f(\M{T}); \q \big]$) performs much better for DSMN* than DSMN becuase of the intermdiate supervision.

Further, Table~\ref{tab:hops} shows the effect of varying the number of memory `hops' for DSMN and DSMN* on FloorPlanQA. The performance of both DSMN and DSMN* increases with the number of `hops'. Note that even the 1-hop DSMN* performs well (better than baselines). Also, note that the difference in performance between 2-hop DSMN* and 3-hop DSMN* is not much. A possible justification for why DSMN* performs well even with fewer memory `hops' is that DSMN* completes some `hops of reasoning' in the visual representation module itself. Suppose one needs to find the location of an object placed in a room, w.r.t. the house. To do so, one first needs to find the location of the room w.r.t. the house, and then the location of the object w.r.t. the room. However, if one has already `sketched' out the location of the object in the house, one can directly fetch it. It is during sketching the object's location that one has completed a `hop of reasoning'. For a sample from FloorPlanQA, we visualize the attention maps in the memory module of 3-hop DMN+ and 3-hop DSMN* in Fig.~\ref{visualization_fig}. To infer the location of room 1's door, DSMN* directly fetches sentence 3,  while DMN+ tries to do so by fetching two sentences (one for the room's door location w.r.t the room and one for the room's location w.r.t the house).

\noindent \textbf{Conclusion:} We have investigated how to use DNNs for modeling visual thinking. We have introduced two synthetic QA datasets, FloorPlanQA and ShapeIntersection, that test a system's ability to think visually. We have developed DSMN, a novel DNN that reasons in the visual space for answering questions. Experimental results have demonstrated the effectiveness of DSMN for geometric reasoning on synthetic data.

\noindent \textbf{Acknowledgements:} This work is partially supported by the National Science Foundation under Grant No. 1633157.

\appendix
\section{Appendix}
\label{sec:appendix}
\begin{table}[!ht]
  \centering
  \begin{center}
    \footnotesize
    \begin{tabular}{l||c|c}
    MODEL  & Embedding size & $l_2$ regularization  \\ 
    \hline \hline
    LSTM-1 & 128 & 1e-4 \\
    LSTM-2 & 512 & 1e-5 \\
    MemN2N & 64  & NA   \\
    DMN+   & 64  & 1e-3 \\
    DSMN   & 32  & 1e-3 \\
    DSMN*  & 32  & 1e-4 \\
    \end{tabular}
  \end{center}
  \caption{The value of the tuned hyper-parameters for all the models on FloorPlanQA. MemN2N~\cite{sukhbaatar2015} model does not use $l_2$ regularization, so we tuned only the embedding size}
  \label{tab:variance}
\end{table}

\begin{table}[!ht]
  \centering
  \begin{center}
    \footnotesize
    \begin{tabular}{l||c|c}
    MODEL  & Embedding size & $l_2$ regularization  \\ 
    \hline \hline
    LSTM-1 & 2048 & 0.01 \\
    LSTM-2 & 512  & 0.1  \\
    MemN2N & 512  & NA   \\
    DMN+   & 128  & 0.1  \\
    DSMN   & 32   & 0.1  \\
    DSMN*  & 64   & 0.01 \\
    \end{tabular}
  \end{center}
  \caption{The value of the tuned hyper-parameters for all the models on ShapeIntersection. MemN2N~\cite{sukhbaatar2015} model does not use $l_2$ regularization, so we tuned only the embedding size.}
  \label{tab:variance}
\end{table}

\bibliography{acl2018}

\begin{thebibliography}{32}
\expandafter\ifx\csname natexlab\endcsname\relax\def\natexlab#1{#1}\fi

\bibitem[{Antol et~al.(2015)Antol, Agrawal, Lu, Mitchell, Batra,
  Lawrence~Zitnick, and Parikh}]{antol2015}
Stanislaw Antol, Aishwarya Agrawal, Jiasen Lu, Margaret Mitchell, Dhruv Batra,
  C~Lawrence~Zitnick, and Devi Parikh. 2015.
\newblock Vqa: Visual question answering.
\newblock In \emph{ICCV}, pages 2425--2433.

\bibitem[{Cho et~al.(2014)Cho, Van~Merri{\"e}nboer, Bahdanau, and
  Bengio}]{cho2014properties}
Kyunghyun Cho, Bart Van~Merri{\"e}nboer, Dzmitry Bahdanau, and Yoshua Bengio.
  2014.
\newblock On the properties of neural machine translation: Encoder-decoder
  approaches.
\newblock \emph{arXiv preprint arXiv:1409.1259}.

\bibitem[{Fitting(2012)}]{fitting2012first}
Melvin Fitting. 2012.
\newblock \emph{First-order logic and automated theorem proving}.
\newblock Springer Science \& Business Media.

\bibitem[{Forbus et~al.(1991)Forbus, Nielsen, and
  Faltings}]{forbus1991qualitative}
Kenneth~D Forbus, Paul Nielsen, and Boi Faltings. 1991.
\newblock Qualitative spatial reasoning: The clock project.
\newblock \emph{Artificial Intelligence}, 51(1-3):417--471.

\bibitem[{Graves et~al.(2014)Graves, Wayne, and Danihelka}]{graves2014neural}
Alex Graves, Greg Wayne, and Ivo Danihelka. 2014.
\newblock Neural turing machines.
\newblock \emph{arXiv preprint arXiv:1410.5401}.

\bibitem[{Graves et~al.(2016)Graves, Wayne, Reynolds, Harley, Danihelka,
  Grabska-Barwi{\'n}ska, Colmenarejo, Grefenstette, Ramalho, Agapiou
  et~al.}]{graves2016hybrid}
Alex Graves, Greg Wayne, Malcolm Reynolds, Tim Harley, Ivo Danihelka, Agnieszka
  Grabska-Barwi{\'n}ska, Sergio~G{\'o}mez Colmenarejo, Edward Grefenstette,
  Tiago Ramalho, John Agapiou, et~al. 2016.
\newblock Hybrid computing using a neural network with dynamic external memory.
\newblock \emph{Nature}, pages 471--476.

\bibitem[{Gregor et~al.(2015)Gregor, Danihelka, Graves, Rezende, and
  Wierstra}]{gregor2015draw}
Karol Gregor, Ivo Danihelka, Alex Graves, Danilo~Jimenez Rezende, and Daan
  Wierstra. 2015.
\newblock Draw: A recurrent neural network for image generation.
\newblock \emph{arXiv preprint arXiv:1502.04623}.

\bibitem[{Gribble et~al.(1998)Gribble, Browning, Hewett, Remolina, and
  Kuipers}]{gribble1998integrating}
William~S Gribble, Robert~L Browning, Micheal Hewett, Emilio Remolina, and
  Benjamin~J Kuipers. 1998.
\newblock Integrating vision and spatial reasoning for assistive navigation.
\newblock In \emph{Assistive Technology and artificial intelligence}, pages
  179--193.

\bibitem[{Gupta et~al.(2017)Gupta, Davidson, Levine, Sukthankar, and
  Malik}]{gupta2017cognitive}
Saurabh Gupta, James Davidson, Sergey Levine, Rahul Sukthankar, and Jitendra
  Malik. 2017.
\newblock Cognitive mapping and planning for visual navigation.
\newblock \emph{arXiv preprint arXiv:1702.03920}.

\bibitem[{Henaff et~al.(2016)Henaff, Weston, Szlam, Bordes, and
  LeCun}]{henaff2016tracking}
Mikael Henaff, Jason Weston, Arthur Szlam, Antoine Bordes, and Yann LeCun.
  2016.
\newblock Tracking the world state with recurrent entity networks.
\newblock \emph{arXiv preprint arXiv:1612.03969}.

\bibitem[{Hochreiter and Schmidhuber(1997)}]{hochreiter1997long}
Sepp Hochreiter and J{\"u}rgen Schmidhuber. 1997.
\newblock Long short-term memory.
\newblock \emph{Neural computation}, pages 1735--1780.

\bibitem[{Johnson et~al.(2016)Johnson, Hariharan, van~der Maaten, Fei-Fei,
  Zitnick, and Girshick}]{johnson2016}
Justin Johnson, Bharath Hariharan, Laurens van~der Maaten, Li~Fei-Fei,
  C~Lawrence Zitnick, and Ross Girshick. 2016.
\newblock Clevr: A diagnostic dataset for compositional language and elementary
  visual reasoning.
\newblock \emph{arXiv preprint arXiv:1612.06890}.

\bibitem[{Kingma and Ba(2014)}]{kingma2014adam}
Diederik Kingma and Jimmy Ba. 2014.
\newblock Adam: A method for stochastic optimization.
\newblock \emph{arXiv preprint arXiv:1412.6980}.

\bibitem[{Kumar et~al.(2016)Kumar, Irsoy, Ondruska, Iyyer, Bradbury, Gulrajani,
  Zhong, Paulus, and Socher}]{kumar2016}
Ankit Kumar, Ozan Irsoy, Peter Ondruska, Mohit Iyyer, James Bradbury, Ishaan
  Gulrajani, Victor Zhong, Romain Paulus, and Richard Socher. 2016.
\newblock Ask me anything: Dynamic memory networks for natural language
  processing.
\newblock In \emph{ICML}, pages 1378--1387.

\bibitem[{Landau and Jackendoff(1993)}]{landau1993whence}
Barbara Landau and Ray Jackendoff. 1993.
\newblock Whence and whither in spatial language and spatial cognition?
\newblock \emph{Behavioral and brain sciences}, 16:255--265.

\bibitem[{Lathrop and Laird(2007)}]{lathrop2007towards}
Scott~D Lathrop and John~E Laird. 2007.
\newblock Towards incorporating visual imagery into a cognitive architecture.
\newblock In \emph{International conference on cognitive modeling}, page~25.

\bibitem[{Levesque et~al.(2011)Levesque, Davis, and
  Morgenstern}]{levesque2011winograd}
Hector~J Levesque, Ernest Davis, and Leora Morgenstern. 2011.
\newblock The winograd schema challenge.
\newblock In \emph{AAAI Spring Symposium}, volume~46, page~47.

\bibitem[{Lin and Parikh(2015)}]{lin2015}
Xiao Lin and Devi Parikh. 2015.
\newblock Don't just listen, use your imagination: Leveraging visual common
  sense for non-visual tasks.
\newblock In \emph{ICCV}, pages 2984--2993.

\bibitem[{Lu et~al.(2016)Lu, Yang, Batra, and Parikh}]{lu2016hierarchical}
Jiasen Lu, Jianwei Yang, Dhruv Batra, and Devi Parikh. 2016.
\newblock Hierarchical question-image co-attention for visual question
  answering.
\newblock In \emph{NIPS}, pages 289--297.

\bibitem[{M{\"u}ller et~al.(2000)M{\"u}ller, R{\"o}fer, Lankenau, Musto, Stein,
  and Eisenkolb}]{muller2000coarse}
Rolf M{\"u}ller, Thomas R{\"o}fer, Axel Lankenau, Alexandra Musto, Klaus Stein,
  and Andreas Eisenkolb. 2000.
\newblock Coarse qualitative descriptions in robot navigation.
\newblock In \emph{Spatial Cognition II}, pages 265--276.

\bibitem[{Perez and Liu(2016)}]{perez2016dialog}
Julien Perez and Fei Liu. 2016.
\newblock Dialog state tracking, a machine reading approach using memory
  network.
\newblock \emph{arXiv preprint arXiv:1606.04052}.

\bibitem[{Pinker(2003)}]{pinker2003language}
Steven Pinker. 2003.
\newblock \emph{The language instinct: How the mind creates language}.
\newblock Penguin UK.

\bibitem[{Reed et~al.(2016)Reed, Akata, Yan, Logeswaran, Schiele, and
  Lee}]{reed2016generative}
Scott Reed, Zeynep Akata, Xinchen Yan, Lajanugen Logeswaran, Bernt Schiele, and
  Honglak Lee. 2016.
\newblock Generative adversarial text to image synthesis.
\newblock \emph{arXiv preprint arXiv:1605.05396}.

\bibitem[{Richardson et~al.(2013)Richardson, Burges, and
  Renshaw}]{richardson2013mctest}
Matthew Richardson, Christopher~JC Burges, and Erin Renshaw. 2013.
\newblock Mctest: A challenge dataset for the open-domain machine comprehension
  of text.
\newblock In \emph{EMNLP}, volume~3, page~4.

\bibitem[{Seo et~al.(2015)Seo, Hajishirzi, Farhadi, Etzioni, and
  Malcolm}]{seo2015}
Minjoon Seo, Hannaneh Hajishirzi, Ali Farhadi, Oren Etzioni, and Clint Malcolm.
  2015.
\newblock Solving geometry problems: Combining text and diagram interpretation.
\newblock In \emph{EMNLP}, pages 1466--1476.

\bibitem[{Shariff(1998)}]{shariff1998natural}
A~Rashid~BM Shariff. 1998.
\newblock Natural-language spatial relations between linear and areal objects:
  the topology and metric of english-language terms.
\newblock \emph{International journal of geographical information science},
  12:215--245.

\bibitem[{Skubic et~al.(2004)Skubic, Perzanowski, Blisard, Schultz, Adams,
  Bugajska, and Brock}]{skubic2004spatial}
Marjorie Skubic, Dennis Perzanowski, Samuel Blisard, Alan Schultz, William
  Adams, Magda Bugajska, and Derek Brock. 2004.
\newblock Spatial language for human-robot dialogs.
\newblock \emph{IEEE Transactions on Systems, Man, and Cybernetics, Part C
  (Applications and Reviews)}, pages 154--167.

\bibitem[{Sukhbaatar et~al.(2015)Sukhbaatar, Weston, Fergus
  et~al.}]{sukhbaatar2015}
Sainbayar Sukhbaatar, Jason Weston, Rob Fergus, et~al. 2015.
\newblock End-to-end memory networks.
\newblock In \emph{NIPS}, pages 2440--2448.

\bibitem[{Vaswani et~al.(2017)Vaswani, Shazeer, Parmar, Uszkoreit, Jones,
  Gomez, Kaiser, and Polosukhin}]{vaswani2017attention}
Ashish Vaswani, Noam Shazeer, Niki Parmar, Jakob Uszkoreit, Llion Jones,
  Aidan~N Gomez, Lukasz Kaiser, and Illia Polosukhin. 2017.
\newblock Attention is all you need.
\newblock \emph{arXiv preprint arXiv:1706.03762}.

\bibitem[{Weston et~al.(2015)Weston, Bordes, Chopra, Rush, van Merri{\"e}nboer,
  Joulin, and Mikolov}]{weston2015}
Jason Weston, Antoine Bordes, Sumit Chopra, Alexander~M Rush, Bart van
  Merri{\"e}nboer, Armand Joulin, and Tomas Mikolov. 2015.
\newblock Towards ai-complete question answering: A set of prerequisite toy
  tasks.
\newblock \emph{arXiv preprint arXiv:1502.05698}.

\bibitem[{Xiong et~al.(2016)Xiong, Merity, and Socher}]{xiong2016}
Caiming Xiong, Stephen Merity, and Richard Socher. 2016.
\newblock Dynamic memory networks for visual and textual question answering.
\newblock In \emph{ICML}, pages 2397--2406.

\bibitem[{Zelek(1997)}]{zelek1997human}
John~S Zelek. 1997.
\newblock Human-robot interaction with minimal spanning natural language
  template for autonomous and tele-operated control.
\newblock In \emph{IROS}, pages 299--305.

\end{thebibliography}
\bibliographystyle{acl_natbib}

\end{document}